%% file: main.tex
\documentclass{article} 
\usepackage{iclr2024_conference,times}
\usepackage{xcolor}
\usepackage{graphicx}
\usepackage{wrapfig}
\usepackage{color}
\usepackage{graphicx}
\usepackage{amssymb,amsmath,amsthm,amsfonts}
\usepackage{mathtools,physics}
\usepackage{caption}
\usepackage{subcaption}
\usepackage{siunitx}
\usepackage{quantikz}
\usepackage[normalem]{ulem}
\usepackage{hyperref}
\usepackage{algorithm}
\usepackage{algpseudocode}
\usepackage{soul}
\usepackage{titlesec}
\titleformat*{\paragraph}{\itshape\bfseries}

\input{math_commands.tex}

\usepackage{hyperref}
\usepackage{url}

\title{Do Diffusion Models Learn Semantically Meaningful and Efficient Representations?}

\author{Qiyao Liang \\
Department of EECS\\
MIT\\
Cambridge, MA 02139, USA \\
\texttt{qiyao@mit.edu} \\
\And
Ziming Liu \\
Department of Physics \\
MIT\\
Cambridge, MA 02139, USA \\
\texttt{zmliu@mit.edu} \\
\AND
Ila R. Fiete \\
Department of BCS \\
MIT\\
Cambridge, MA 02139, USA \\
\texttt{fiete@mit.edu} \\
}

\iclrfinalcopy 
\begin{document}

\maketitle

\begin{abstract}

Diffusion models have shown the ability to generate images with unconventional compositions, such as astronauts riding horses on the moon, indicating compositional generalization. However, the vast size and the complex nature of realistic datasets make it challenging to quantitatively probe diffusion model's abilility to compositionally generalize. Here, we consider a highly reduced setting to examine whether diffusion models learn semantically meaningful and fully factorized representations of composable features. We perform controlled experiments on conditional DDPMs learning to generate 2D spherical Gaussian bumps centered at specified $x$- and $y$-positions. En route to successful learning semantically meaningful representations, we observe three distinct learning \textit{phases}:
(\textit{phase A}) no latent structure, (\textit{phase B}) a 2D manifold of disordered states, and (\textit{phase C}) a 2D ordered manifold, each with distinct generation behavior/failure modes. Furthermore, we show that even under imbalanced datasets where features ($x$- versus $y$-positions) are represented with skewed frequencies, the learning process for $x$ and $y$ is coupled rather than factorized, potentially indicating that the model does not learn a fully factorized hence efficient representation.  

\end{abstract}
\section{Introduction}

\subsection{Background}

Text-to-image generative models have demonstrated incredible ability in generating photo-realistic images that involve combining elements in innovative ways that are not present in the training dataset (e.g. astronauts riding a horse on the moon)~\citep{saharia2022photorealistic,rombach2022high,ramesh2021zero}. A naïve possibility is that the training dataset contains all possible combinations of all elements, and the model memorizes all of these. This would require massive amounts of data, given that the number of such combinations grows exponentially with the number of elements. The success of generative models at constructing improbable combinations of elements suggests that they are able to compositionally generalize, by learning factorized internal representations of individual elements, and then composing those representations in new ways ~\citep{ du2021, yang2023, du2023}. However, given the massive datasets on which at-scale generative models are trained, it is difficult to quantitatively assess their ability to extract and combine independent elements in the input datasets. The question we would like to answer is how well diffusion models learn semantically meaningful and factorized representations.

\begin{figure}[h]
\begin{center}
\includegraphics[width=4.0in]{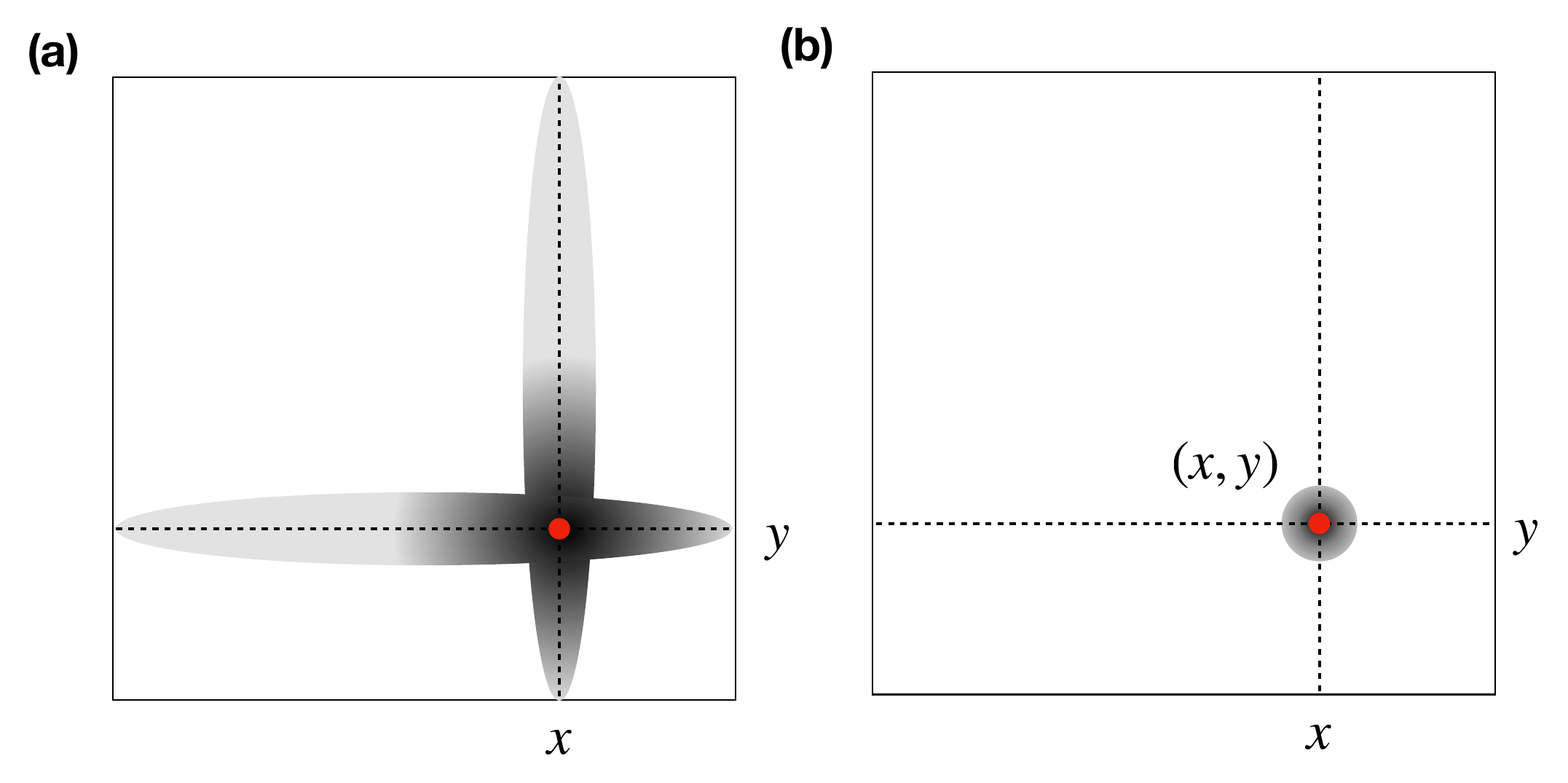}
\end{center}
\setlength{\belowcaptionskip}{-10pt}
\caption{\textbf{Schematic illustration of a fully factorized solution vs. an unfactorized solution.} \textbf{(a)} is a factorized solution where the $x$ position is generated independently from
the $y$ position of the Gaussian bump by intersecting two oval Gaussian bumps localized in one dimension but not the other. \textbf{(b)} shows a coupled solution where a single Gaussian bump localized in both dimension is generated. One difference between the two possibilities is that a network that recognized the independence of generation in $x,y$ could learn with $\mathcal{O}(2K)$ examples, while otherwise it would  take $\mathcal{O}(K^2)$ examples.}
\label{fig:schematic}
\end{figure}

To answer this question, we propose a simple task, which is to reconstruct an image with a 2D spherical Gaussian bump centered at various, independently varying $x$ and $y$ locations. A naive solution is to memorize all possible combinations of the $x$ and $y$ locations, which is expensive. Alternatively, the model can learn $x$ and $y$ as factorized 1D concepts and combine them compositionally. A schematic illustration of the two solutions are depicted in Fig.~\ref{fig:schematic}. Which solution will the model learn?

Specifically we conduct controlled experiments in this setting to investigate the following questions:
\begin{enumerate}
    \item How does the representation learned by the model relate to its performance?
    \item How and under what conditions do semantically meaningful representations emerge? How does training data affect the model's learned representation?
    \item Are the learned representations of the models factorized under imbalanced datasets?
\end{enumerate}

\subsection{Our Contributions}
In this work, we aim to tackle the questions posed above via an empirical study of a toy conditional diffusion model using synthetic datasets that can be controllably varied. Our key findings can be summarized as follows:
\begin{itemize}
    \item {\bf Diffusion models undergo three learning phases.} We observe the three phases of manifold formation, including three distinct failure modes along the training progress.
    \item {\bf  Performance is highly correlated with the learned representation.} We find that the formation of an ordered manifold is a strong indicator of good model performance.
    \item {\bf Diffusion models learn semantically meaningful representations.} In the terminal learning phase, a semantically meaningful representation emerges, and the rate at which it emerges depends on the data density. 
    \item {\bf The learning process of independent concepts is not fully factorized.}  We discover that even in imbalanced datasets with skewed representation of independent concepts, the learning process is not fully factorized, indicating that the model might not have learned a fully factorized, efficient representation.
\end{itemize}

\section{Experimental Setup}

\textbf{Dataset Generation. }
We generate 32 $\times$ 32 grayscale images each consisting of a single 2D spherical Gaussian bump at various locations of the image. 
\begin{wrapfigure}{r}{0.25\textwidth}
\begin{center}
\includegraphics[width=0.25\textwidth]{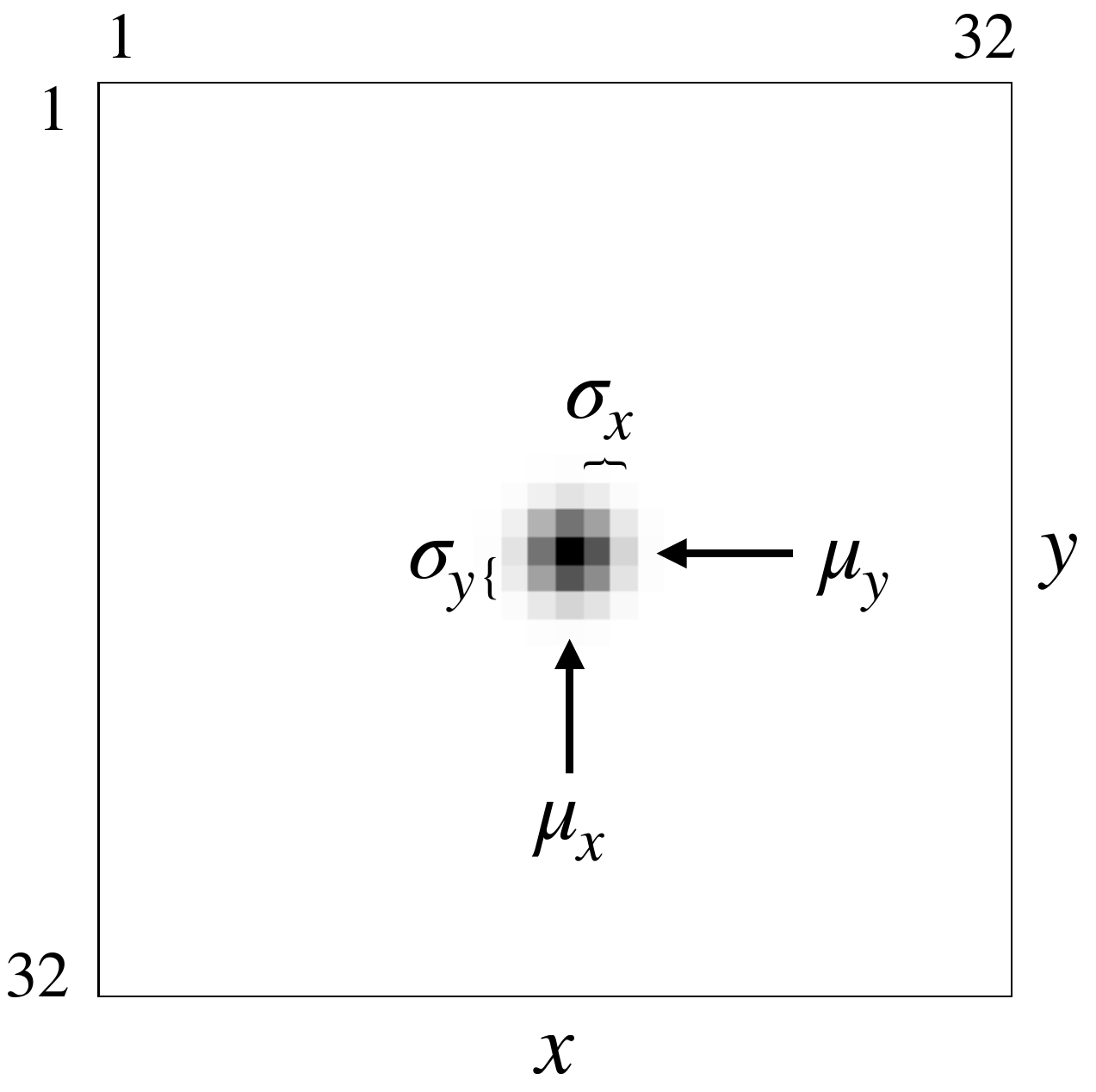}
\end{center}
\caption{Example image data.}
\label{fig:sample_image}
\end{wrapfigure}
The brightness of the pixel $v_{(x,y)}$ at position $(x,y)$ in a given image with a 2D Gaussian bump centered at $(\mu_x, \mu_y)$ with standard deviations of $(\sigma_x, \sigma_y)$ is given by $v_{(x,y)} = 255 \times (1-\exp{-(x-\mu_x)^2/4\sigma_x^2-(y-\mu_y)^2/4\sigma_y^2})$ with the normalized range of $v_{(x,y)}$ to be $[0,255]$. Each image is generated with a ground truth label of $(\mu_x, \mu_y)$, which continuously vary within $[0,32]^2$. In our convention of notation, we label the top left corner of the image as $(1,1)$ while the bottom right corner of the image as $(32,32)$. A sample image centered at $\mu_x=\mu_y=16$ with $\sigma_x=\sigma_y=1$ is shown in Fig.~\ref{fig:sample_image}.

A dataset of these images consist of the enumeration of all possible Gaussians tiling the whole 32 $\times$ 32 canvas at increment $d_x$ in the $x$-direction and $d_y$ in the $y$-direction. A larger $d_x$ or $d_y$ means a sparser tiling of the image space and less abundant data while a smaller $d_x$ or $d_y$ result in more abundant data with denser tiling of the total image space. In a single dataset, we assume the spread of the Gaussian bump $\sigma_x=\sigma_y:=\sigma$ to be constant. With a larger spread leading to more spatial information of neighboring Gaussian bump and a smaller spread less information. By parameterically tuning the increments $d_x$ and $d_y$ and the spread $\sigma$, we can generate datasets of various sparsities and overlaps. We provide some a more detailed analysis of the various attributes of the data based on these parameters in Appendix Sec.~\ref{sec:app_dataset}.

\textbf{Models. } 
We train a conditional Denoising Diffusion Probabilistic Models (DDPM)~\citep{ho2020denoising,cond1,cond2} with a standard UNet architecture as shown in Fig.~\ref{fig:architecture} in the Appendix. For each image in the training dataset, we provide an explicit ground truth label $(\mu_x,\mu_y)$ as the input to the embedding. For reference, we investigate the internal representation learned by the model using the output of layer 4 as labeled in Fig.~\ref{fig:architecture}. Since each dataset has inherently two latent dimensions, $x$- and $y$-positions, we use dimension reduction tools to reduce the internal representation of the model to a 3D embedding for the ease of visualization and analysis.
We defer the details of model architecture, dimension reduction, and experimentation to Appendix Sections~\ref{sec:app_architecture} and ~\ref{sec:app_umap}.

\begin{figure}[!htb]
\begin{center}
\includegraphics[width=4.8in]{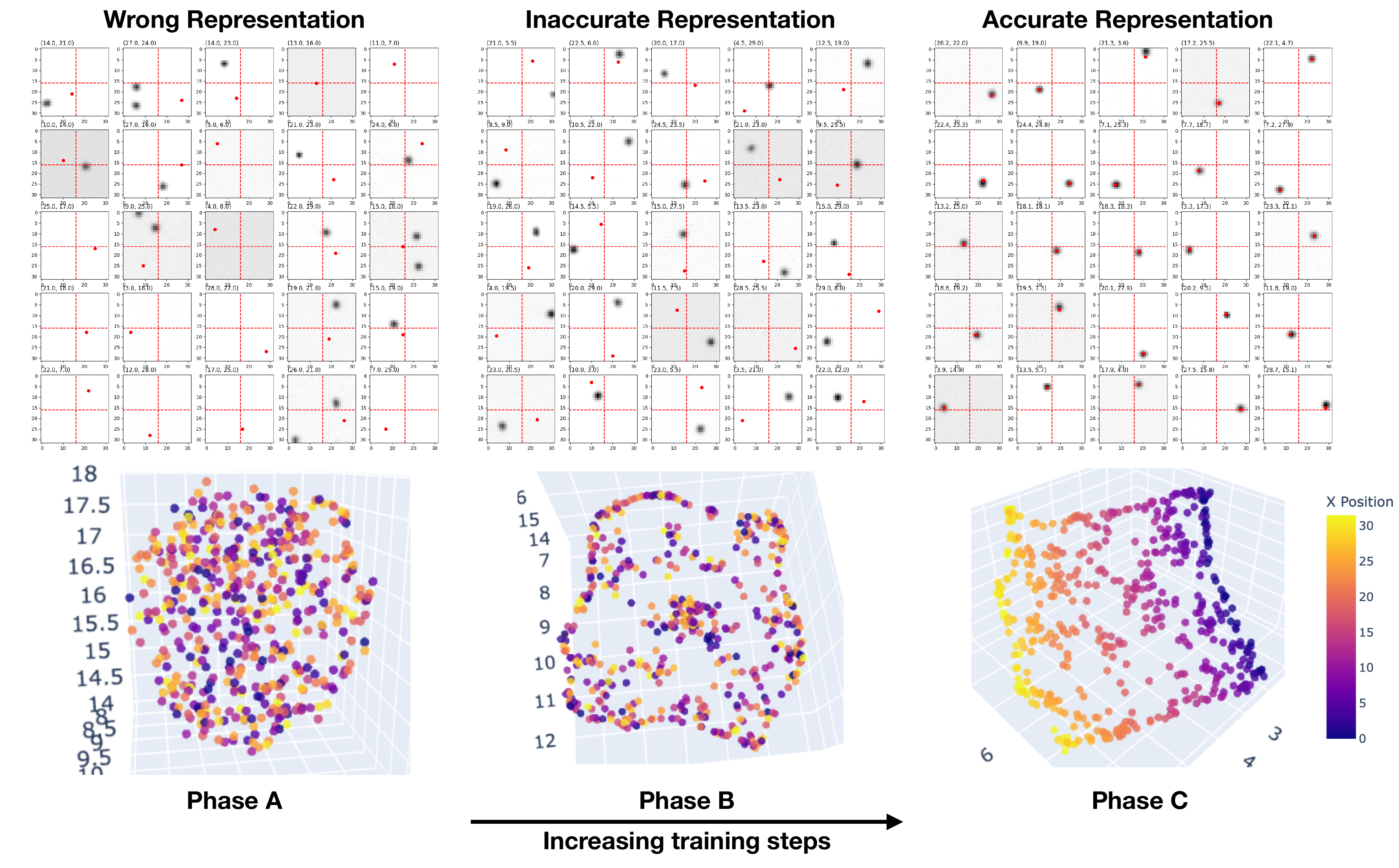}
\end{center}
\caption{\textbf{The three phases of manifold formation.} The learned representations (UMAP reduced, colored by the ground truth $x$-positions) of the diffusion models undergo the three phases in increasing order of training steps as depicted in the 3D visualizations in the bottom row. In each phase, the corresponding qualitative generation behavior is demonstrated with 25 sampled images in the top row, in which the red dots mark the ground truth locations of the Gaussian bumps. (\textit{Phase A}) has no particular structure in the learned representation, and the generated images either have no Gaussian bumps or multiple Gaussian bumps at the wrong locations. (\textit{Phase B}) has a disordered, quasi-2D manifold with corresponding generation behavior of a single Gaussian bump at the wrong location. (\textit{Phase C}) has an ordered 2D manifold with the desired generation behavior. }
\label{fig:three_phases}
\end{figure}

\textbf{Evaluations. }
To briefly summarize, we assess the performance of the model based on the accuracies of the images generated and the quality of fit of the 3D embedding of the internal representation corresponding to the sampled images in predicting the ground truth image labels. We refer to the two quantitative performance indicators as the \textit{predicted label accuracy} and the \textit{averaged R-squared}. Intuitively, these two metrics range from 0 to 1, with the closer they are to 1 the higher quality of the generated images/learned representation, i.e., the better the performance of the model. Further details on these metrics can be found in Appendix Sec.~\ref{sec:app_metrics}.

\section{Results}
\label{sec:results}

{\bf Three phases in training.} We train various diffusion models on datasets of various increments $d$ and $\sigma$ between the value of 0.1 to 1.0. For a fair comparison across these models, we fix the total amount of training steps for all models, as measured in units of batches (see Appendix Sec.~\ref{sec:app_training}). As we increase the amount of training for a given model, we observe the universal emergence of three phases in manifold formation each corresponding to distinct generation behavior, as shown in Fig.~\ref{fig:three_phases}. In particular, we noted three distinct failure modes during generation, namely i) no Gaussian bump is generated, ii) a single Gaussian bump is generated at an inaccurate location, and iii) multiple Gaussian bumps are generated at inaccurate locations. During the initial phase (phase A), the formed manifold does not have a particular structure or order. The generation behavior during this phase include all three of the above-mentioned failure modes. As we progressively increase the amount of training, we begin to witness phase B emerging, where the manifold formed is 2-dimensional or quasi-2D but unordered. The predominant failure mode of generation during this phase is ii, while the difference between the locations of the generated Gaussian bumps from their ground truths gradually diminishes as we proceed in training. Eventually, the model learns a 2D ordered manifold with the desired generation behavior, reaching the terminal phase C.

\begin{figure}[!htb]
\begin{center}
\includegraphics[width=4.2in]{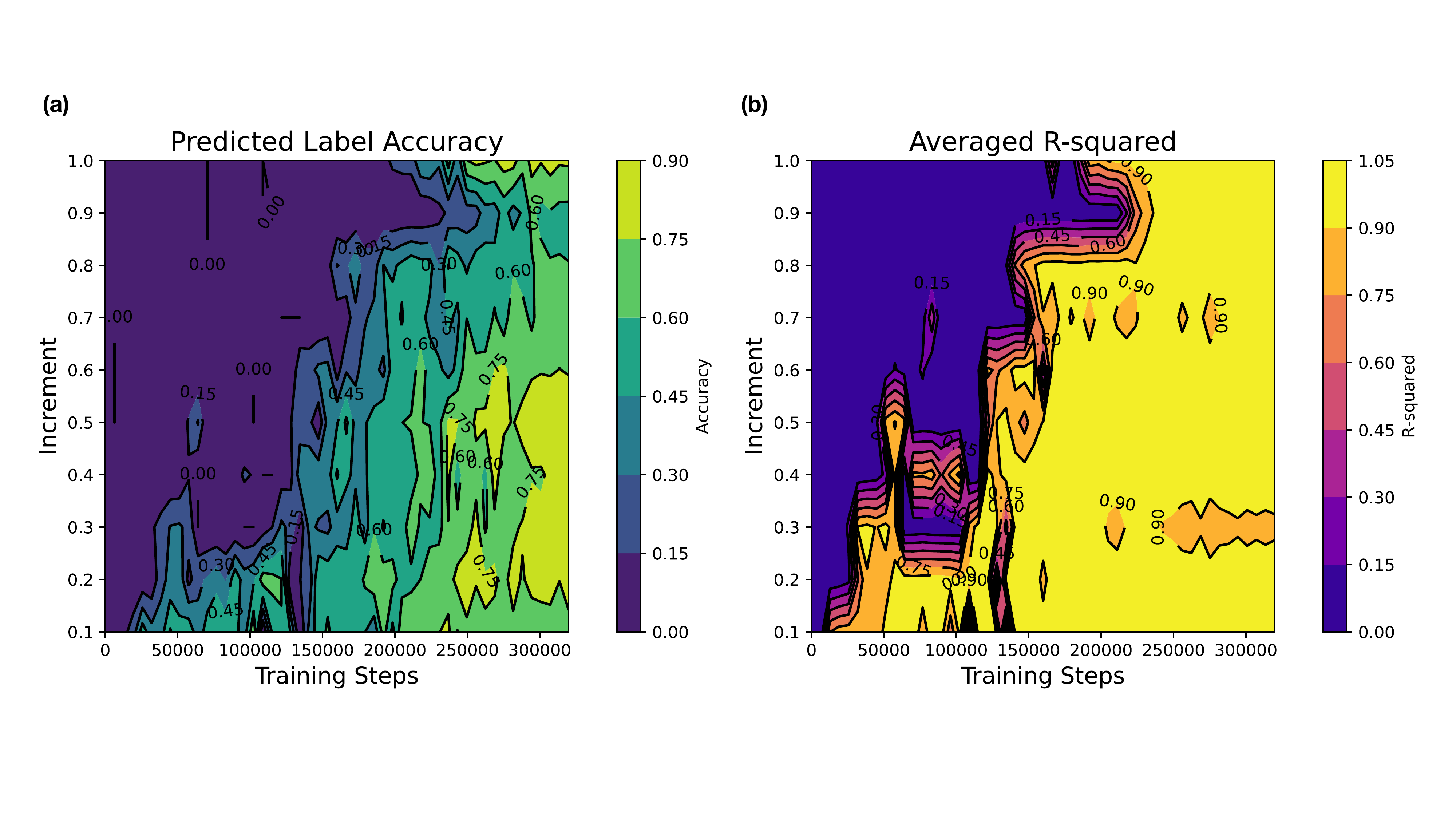}
\end{center}
\setlength{\belowcaptionskip}{-10pt}
\caption{\textbf{2D phase diagram of performance metrics as a function of increment and training steps. } \textbf{(a)} shows the predicted label accuracy and \textbf{(b)} shows the R-squared averaged in predicting $x$- and $y$-positions of the Gaussian bumps from the latent representation. The models are trained with datasets of various increments from 0.1 to 1.0 and sigma of 1.0. The total number of training steps are held constant across all the models.}
\label{fig:phase_increment}
\end{figure}

{\bf Internal representation is key to performance.} To investigate models' performance and representations learned under various datasets, we plot the two performance metrics, accuracy and averaged R-squared, as a function of increasing increments and training steps in Fig.~\ref{fig:phase_increment}. For all datasets of varying increments, we have fixed the spread of the Gaussian bumps to be $\sigma=1.0$. Noticeably, comparing Fig.~\ref{fig:phase_increment} (a) and (b) shows that learning a high-quality representation is key to achieving better accuracies in image generation. Moreover, in general, we observe that datasets with smaller increments lead to faster learning of the desired representation. Hence, given the same amount of training (having seen the same amount of data), the models trained using datasets that are more information-dense will result in a better-quality representation learned. On the other hand, with fewer data, an accurate representation can eventually be learned given enough training to compensate. We briefly comment on the information density of the datasets and discuss the trade-off between overlaps of neighboring Gaussian bumps and sensitivity to the spatial information encoded in Appendix Sec.~\ref{sec:app_dataset}. 


\begin{figure}[h]
\begin{center}
\includegraphics[width=4.8in]{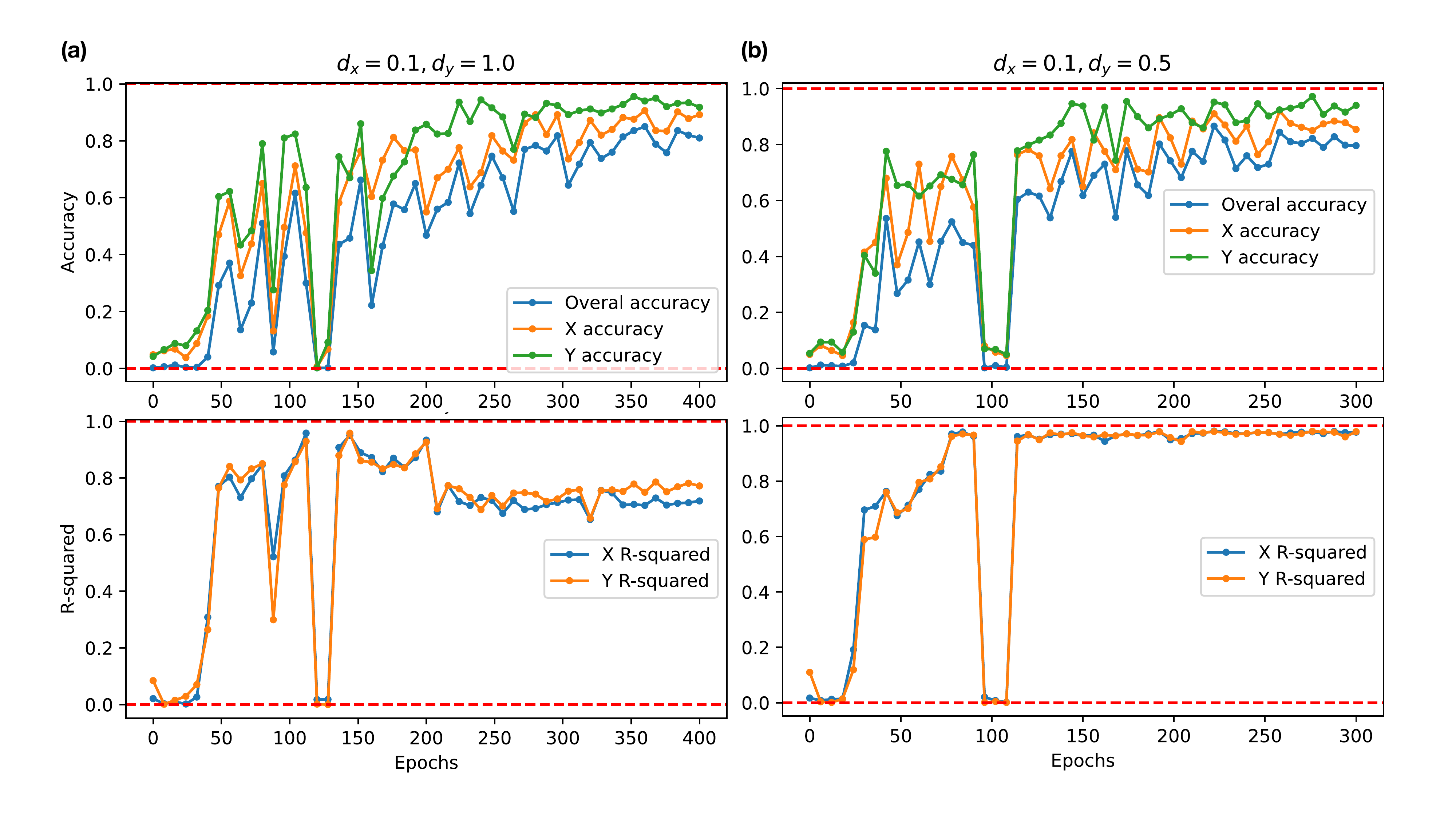}
\end{center}
\setlength{\belowcaptionskip}{-10pt}
\caption{\textbf{Performance metrics of models trained using imbalanced datasets.} \textbf{(a)} using increments of $d_x=0.1$ and $d_y=1.0$ and \textbf{(b)} using increments of $d_x=0.1$ and $d_y=0.5$. Models in both cases are trained with amply sufficient amount of steps to reach convergence. }
\label{fig:imbalanced_exps}
\end{figure}

{\bf The learning rates of $x$ and $y$ are coupled.} Finally, we attempt at answering the question of whether the representation learned is factorized via the investigation of the learning rates of $x$ and $y$. To test that, we train models on datasets that have imbalanced increments in the $x$-direction $d_x$ compared to the y-direction $d_y$. Given such a dataset, we would expect that the model learn these independent concepts at different rates based on the conclusion from Fig.~\ref{fig:phase_increment}, resulting in a factorized manifold. We tested two scenarios, one of stronger imbalance $d_x=0.1$ and $d_y=1.0$, and one of weaker imbalance $d_x=0.1$ and $d_y=0.5$. The performance metrics of the experiments in both cases are shown in Fig.~\ref{fig:imbalanced_exps}(a) and (b), respectively. We see from the figures that despite having more data with finer-grained information of the $x$-positions, the accuracy of generating Gaussian bumps at the correct $y$ locations is generally higher than that at generating at the correct $x$ locations. Moreover, the R-squared values in fitting to the $x$- and the $y$-positions are strongly coupled, which could be indicative that the representations learned are coupled rather than factorized. Overall, we observe that an imbalance in the dataset leads to a deterioration in the general performance of the model rather than factorizing the independent concepts.

\section{Conclusion}


Do diffusion models learn semantically meaningful and efficient representations? We conduct a well-controlled toy model study for diffusion models to learn to generate 2D Gaussian bumps at various $x$- and $y$-positions, given datasets that are parametrically generated to have various densities and overlaps. Throughout the learning process, we observe three phases of the manifold formation and identify corresponding generation behavior with distinctive failure modes in each phase. By comparing models trained under datasets of different sizes and overlaps, we conclude that learning a semantically meaningful representation is essential to model's performance. Moreover, we observed that models learn independent latent features in a coupled fashion despite trained using imbalanced datasets. This potentially indicates that the model learns a coupled representation, which is not as efficient. Future investigation should look into the relation between learning and generation dynamics of diffusion models and their abilities to learn fully factorized representations and compositionally generalize.



\subsubsection*{Acknowledgments}
We thank Yilun Du for helpful discussions at the preliminary stage of our work.

\bibliography{bib}
\bibliographystyle{iclr2024_conference}

\appendix
\section{Related Work}

Compositional generalization has been empirically investigated in many deep generative models before ~\citep{zhao2018, xu2022, okawa2023}. Specifically, \cite{zhao2018} investigated how inductive biases in GANs and VAEs affect their ability to compositionally generalize on toy cognitive psychological tasks. Similarly, \cite{xu2022} developed an evaluation protocol to assess the performance of several unsupervised representation learning algorithms in compositional generalization, where they discovered that disentangled representations do not guarantee better generalization. In a recent empirical study of toy diffusion models,~\cite{okawa2023} shows that diffusion models learn to compositionally generalize multiplicatively. They, however, did not focus on the mechanistic aspect of the learning dynamics or analyze the representations learned by the models. 

One alternative direction is engineering inductive biases that encourage the emergence of compositionality in diffusion models and beyond~\citep{esmaeili2019, higgins2018, du2021, yang2023, du2023}. \cite{yang2023} applied disentangled representation learning techniques to diffusion models to automatically discover concepts and disentangle the gradient field of DDPMs into sub-gradient fields conditioned in the discovered factors. Along a similar line, \cite{du2021} proposed an unsupervised scheme for discovering and representing concepts as separate energy functions that enables explicit composition and permutation of those concepts. In a series of follow-up works, \cite{liu2021, liu2022, liu2023} explored compositional generation with composable diffusion and energy models, as well as concept discovery in text-to-image models.  
\section{Experimental Details}

\subsection{Architecture}
\label{sec:app_architecture}

\begin{figure}[h]
\begin{center}
\includegraphics[width=5.0in]{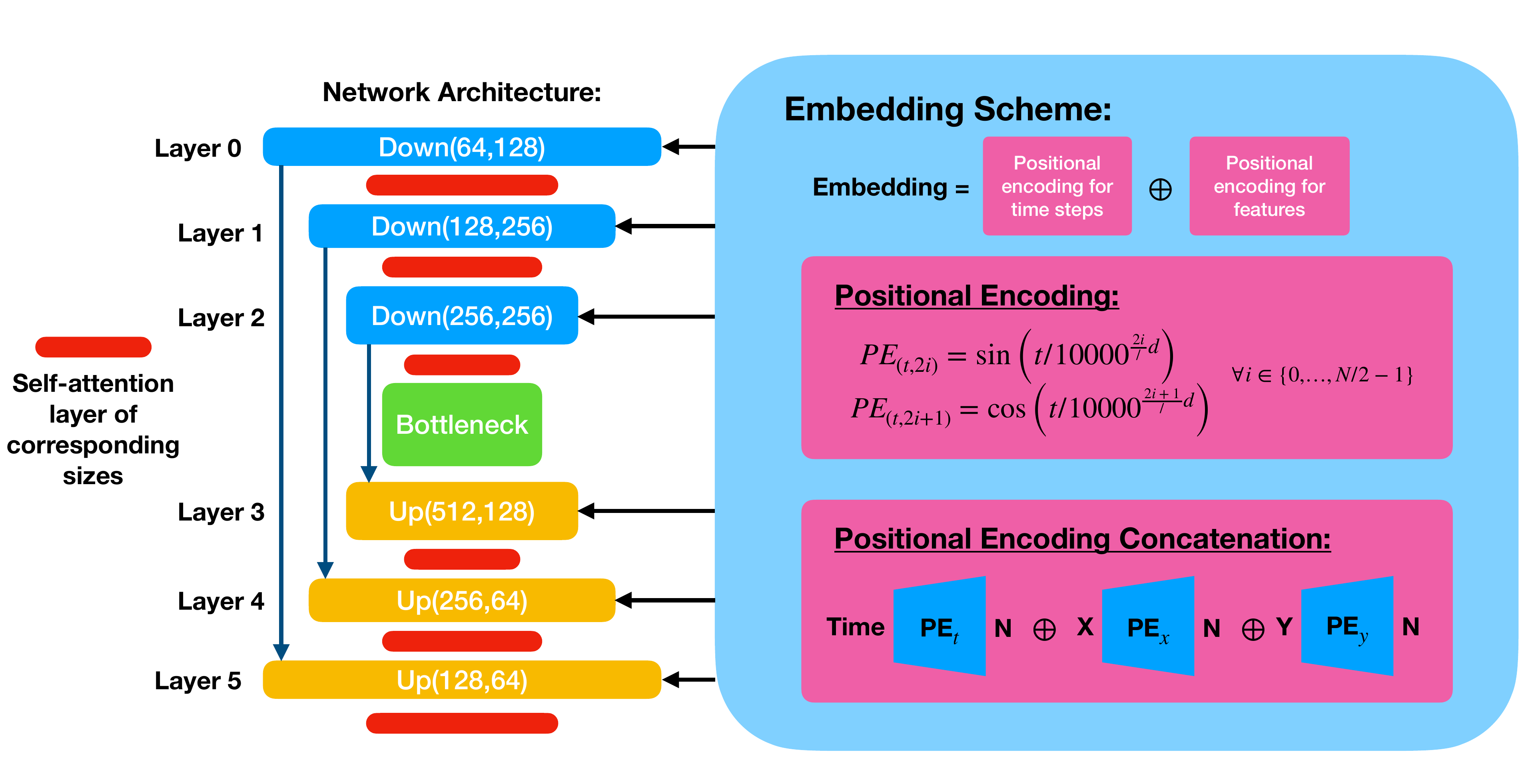}
\end{center}
\caption{\textbf{The UNet architecture of the conditional diffusion model.} The schematic diagram of the standard UNet architecture consisting of three downsampling/upsampling blocks with interlaced self-attention layers and skip connections is shown on the left. The conditional information consisting of a concatenation of positional encodings of timestep and $x$/$y$-positions is passed in at each block as shown on the right.}
\label{fig:architecture}
\end{figure}

We train a conditional Denoising Diffusion Probabilistic Models (DDPM)~\cite{ho2020denoising} with a standard UNet architecture of 3 downsampling and upsampling blocks, interlaced self-attention layers, and skip connections as shown in Fig.~\ref{fig:architecture}. Each down/up-sampling blocks consist of max pooling/upsampling layers followed by two double convolutional layers made up by convolutional layers, group normalization, and GELU activation functions. 

The conditional information is passed in at each down/up-sampling block as shown in the schematic drawing. In our experiments, we choose to preserve the continuity of the Gaussian bump position labels passed into the model via positional encoding rather than using a separate trainable embedding layer at each block. Each embedding vector is made by concatenating equal length vectors of the positional encodings of the timestep, the $x$-position, and the $y$-position. 

In our experiments, we visualize the outputs of layer 4 as the internal representation of the diffusion model. We have chosen not to use the output of the bottleneck layer for our study of the learned latent manifold, as we have observed that the bottleneck layers have diminishing signals in most of our experiments. 

\subsection{Dimension Reduction}
\label{sec:app_umap}
We primarily use the dimension reduction technique Uniform Manifold Approximation and Projection for Dimension Reduction (UMAP)~\cite{mcinnes2020umap} to study and visualize the learned representation of the model. Specifically, we collect image samples and their corresponding internal representations (outputs of layer 4 from the architecture described in Sec.~\ref{sec:app_architecture}). We then transform the high-dimensional internal representations into a 3D embedding as a sample of the learned representation, which we visualize and analyze. For an implementation of UMAP, we used the Python package~\cite{mcinnes2018umap-software}.

\subsection{Evaluation}
\label{sec:app_metrics}
We assess the performance of the model using two primary criteria: 1) the quality of the denoised images and 2) the quality of the learned representation. 

At a given time during or after training, we generate 500 denoised images and their corresponding internal representations of randomly sampled labels based on the training dataset. We predict the labels from the image generated based on the $x$- and $y$-positions of the generated Gaussian bump in the image. We then compute the accuracy of predicted labels from the ground truth labels averaged over 500 samples as 
\begin{align}
    \text{Accuracy} = \frac{1}{500}\sum_{i=1}^{500} \boldsymbol {1}(|\mu^i_x-\hat{\mu}^i_x|<1)\cdot \boldsymbol {1}(|\mu^i_y-\hat{\mu}^i_y|<1),
\end{align}
where $\boldsymbol {1}(\cdot)$ is an indicator function that returns 1 if the expression within holds true, 0 otherwise. Similarly, we can modify this expression to only assess the accuracy of generated $x$-positions or $y$-positions separately. Here we estimate the center of the Gaussian bump $\hat{\mu}^i_x$ and $\hat{\mu}^i_y$ using a label prediction algorithm described in Alg.~\ref{alg:label_prediction} implemented using Otsu's image thresholding and the contour-finding algorithm in the OpenCV package, abbreviated as \verb|cv2|. In the cases where there are no bumps or more than one bump, the algorithm defaults back to finding the centroid of the image. 

\begin{algorithm}
\caption{Label prediction algorithm}
\label{alg:label_prediction} 
\begin{algorithmic}[1]
\Function{LabelPrediction}{$\text{img}$}
    
    \State $(T, \text{thresh}) \gets \text{cv2.threshold}(\text{img}, 0, 255, \text{cv2.THRESH\_BINARY\_INV} \, | \, \text{cv2.THRESH\_OTSU})$
    
    \State $\text{contours}, \_ \gets \text{cv2.findContours}(\text{thresh}, \text{cv2.RETR\_ExTERNAL}, \text{cv2.CHAIN\_APPROx\_SIMPLE})$
    
    \If{$\text{len(contours)} = 0$ \textbf{or} $\text{len(contours)} > 1$}
        \State \Return \textsc{ComputeCentroid2D}($\text{img}$)
    \EndIf
    
    \State $\text{max\_contour} \gets \text{max(contours, key=cv2.contourArea)}$
    
    \State $(x, y, w, h) \gets \text{cv2.boundingRect}(\text{max\_contour})$
    
    \State $cx \gets x + \frac{w}{2}$
    \State $cY \gets y + \frac{h}{2}$
    
    \State \Return $(cx, cY)$
\EndFunction
\end{algorithmic}
\end{algorithm}

To assess the quality of the learned representation, we perform two 1D linear regressions on the UMAP-reduced 3D embedding of the internal representations (outputs of layer 4) corresponding to the 500 sampled images. We use the R-squared values of fit in both predicting $\mu_x$ and $\mu_y$ as an indicator for the quality of the manifold learned in representing $x$- and $y$-positions of the Gaussian bumps.

\subsection{Training Loss}
Diffusion models iteratively denoise a Gaussian noisy image $\mathbf{x}_T$ into a noisefree image $\mathbf{x}_0$ over diffusion timesteps $t\in\{0,1,\ldots, T\}$ given the forward distribution $q(\mathbf{x}_t|\mathbf{x}_{t-1})$ by learning the reverse distribution $p_{\theta}(\mathbf{x}_{t-1}|\mathbf{x}_t)$. Given a conditional cue $\mathbf{c}$, a conditional diffusion model~\citep{cond1,cond2} reconstructs an image from a source distribution $q(\mathbf{x}_0|\mathbf{c})$. Specifically, we train our neural network (UNet) to predict the denoising direction $\epsilon_{\theta}(\mathbf{x}_t, t, \mathbf{c})$ at a given timestep $t$ with conditional cue $\mathbf{c}$ with the goal of minimizing the mean squared loss (MSE) between the predicted and the ground truth noise $\epsilon$ as follows
\begin{align}
    \mathcal{L}:=\mathbb{E}_{t\in\{0,\ldots, T\}, \mathbf{x}_0\sim q(x_0|\mathbf{c}), \epsilon\sim\mathcal{N}(\mathbf{0}, \mathbf{I})}\left[\lVert \epsilon - \epsilon_{\theta}(\mathbf{x}_0, t, \mathbf{c})\rVert^2\right],
\end{align}
where we assume each noise vector $\epsilon$ to be sampled from a Gaussian distribution $\mathcal{N}(\mathbf{0}, \mathbf{I})$ I.I.D. at each timestep $t$.

\subsection{Datasets}
\label{sec:app_dataset}

\begin{figure}[h]
\begin{center}
\includegraphics[width=5.0in]{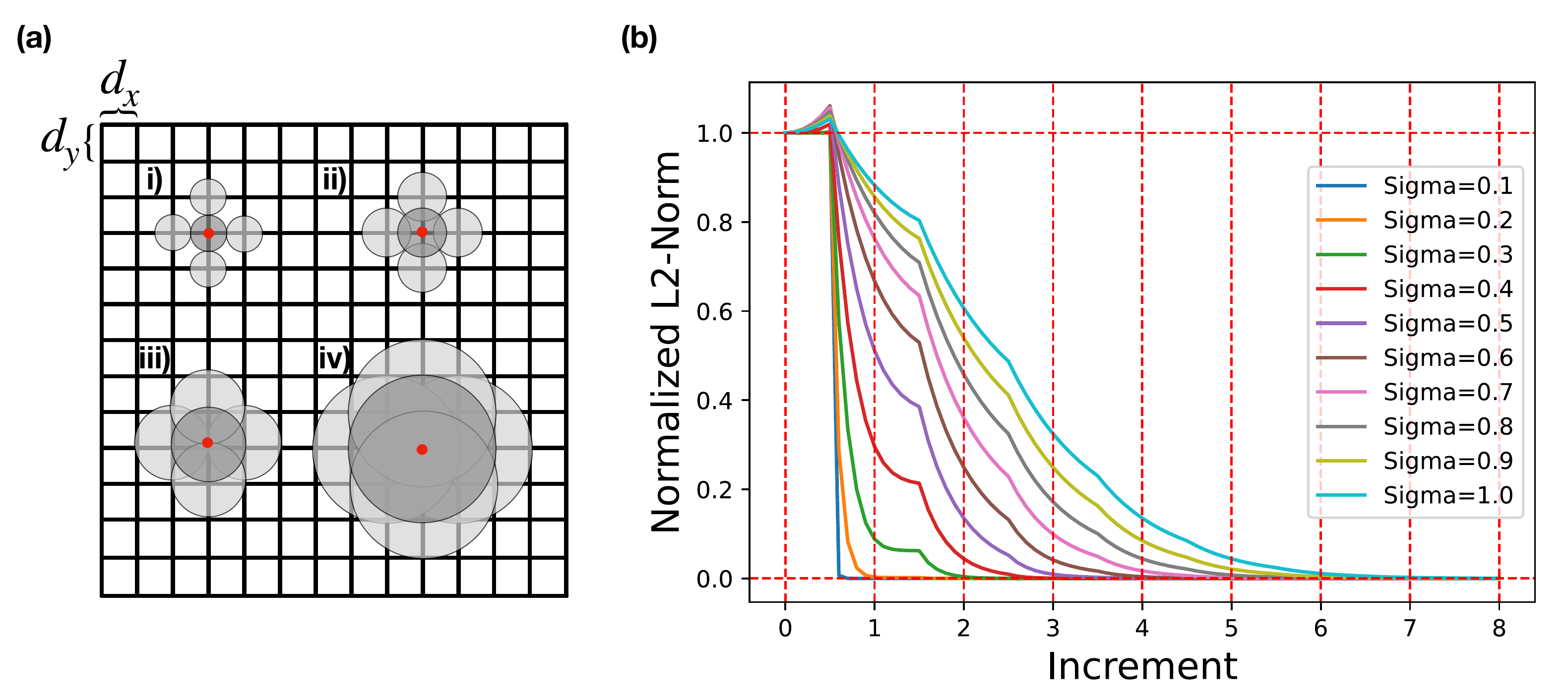}
\end{center}
\caption{\textbf{The overlap of Gaussian bumps with various spreads.} \textbf{(a)} is a schematic demonstration the interplay between the increment $d_x$/$d_y$ and the spread of the Gaussians determined by $\sigma$. The labeled diagram i-iv) shows examples of having no overlap (no neighbor information), nearest neighbor overlap, nearest neighbor and adjacent overlap, and N-nearest neighbors overlap, respectively. \textbf{(b)} shows the normalized L2-norm of the inverted Gaussian image overlaps.}
\label{fig:data_overlap}
\end{figure}

The datasets we used for training the models generating the results in Sec.~\ref{sec:results} have various increments $d_x$/$d_y$ and $\sigma$. Here we briefly comment on the interplay between increments and sigmas, and how they affect dataset densities and overlaps. The ultimate goal of our task of interest is to learn a continuous 2D manifold of all possible locations of the Gaussian bumps. Our datasets are discrete approximations of continuous manifold that can be thought of as a ``web" with each data point as a ``knot," a schematic illustration is shown in Fig.~\ref{fig:data_overlap}(a). Intuitively, the spatial information necessary for an organized, continuous, and semantically meaningful representation to emerge is encoded in the overlap of the neighboring Gaussian bumps, which is tuned via the parameters $d$ and $\sigma$. As we increase $d$, the size of the dataset gets scaled quadratically, resulting in each Gaussian bump to have more overlap with its neighbors.

As we scale up $\sigma$, the dataset size remains fixed while the overlaps with neighbors are significantly increased. In Fig.~\ref{fig:data_overlap}(b), we plot the normalized L2-norm of the product image of neighboring Gaussian bumps as a function of increments for various spreads. Specifically, given two inverted grayscale Gaussian bump images, $a$ and $b$, the normalized L2-norm of their product is given by the formula $\lVert\sqrt{a * b}\rVert_2/\lVert a\rVert_2$, where $*$ is element-wise multiplication and $\lVert\cdot\rVert_2$ is the L2-norm. This quantity should give a rough measure of the image overlap with the exception at increment around 0.5 due to the discrete nature of our data. Moreover, we note that the cusps in the curves occur for the same reason. As we can see, the number of neighbors that a given Gaussian bump has non-trivial overlaps with grows roughly linearly to sub-linearly with the spread. Nonetheless, in Sec.~\ref{sec:app_role_of_sigma} we show that there is no strong correlation between performance or the rate at which semantically meaningful representations emerges and the spread of the Gaussian bumps.

\subsection{Training Details}
\label{sec:app_training}
We train various diffusion models on datasets of various increments $d$ and $\sigma$ from the range $\{0.1,0.2,0.3,0.4,0.5,0.6,0.7,0.8,0.9,1.0\}$. For each model, we fix the total training steps to be 320,000 (number of epochs $\times$ the dataset size in units of batches). We train the models on a quad-core Nvidia A100 GPU, and an average training session lasts around 6 hours. For each model we run three separate seeds and select the run that achieve the optimal accuracy at terminal epoch. To produce the results shown in Fig.~\ref{fig:phase_increment} and Fig.~\ref{fig:phase_sigma}, we sample 500 images as well as their corresponding outputs at layer 4 every 6400 training steps and at the terminal step. 

\section{Additional Results}

\subsection{The role of Sigma}
\label{sec:app_role_of_sigma}



\begin{figure}[h]
\begin{center}
\includegraphics[width=5.0in]{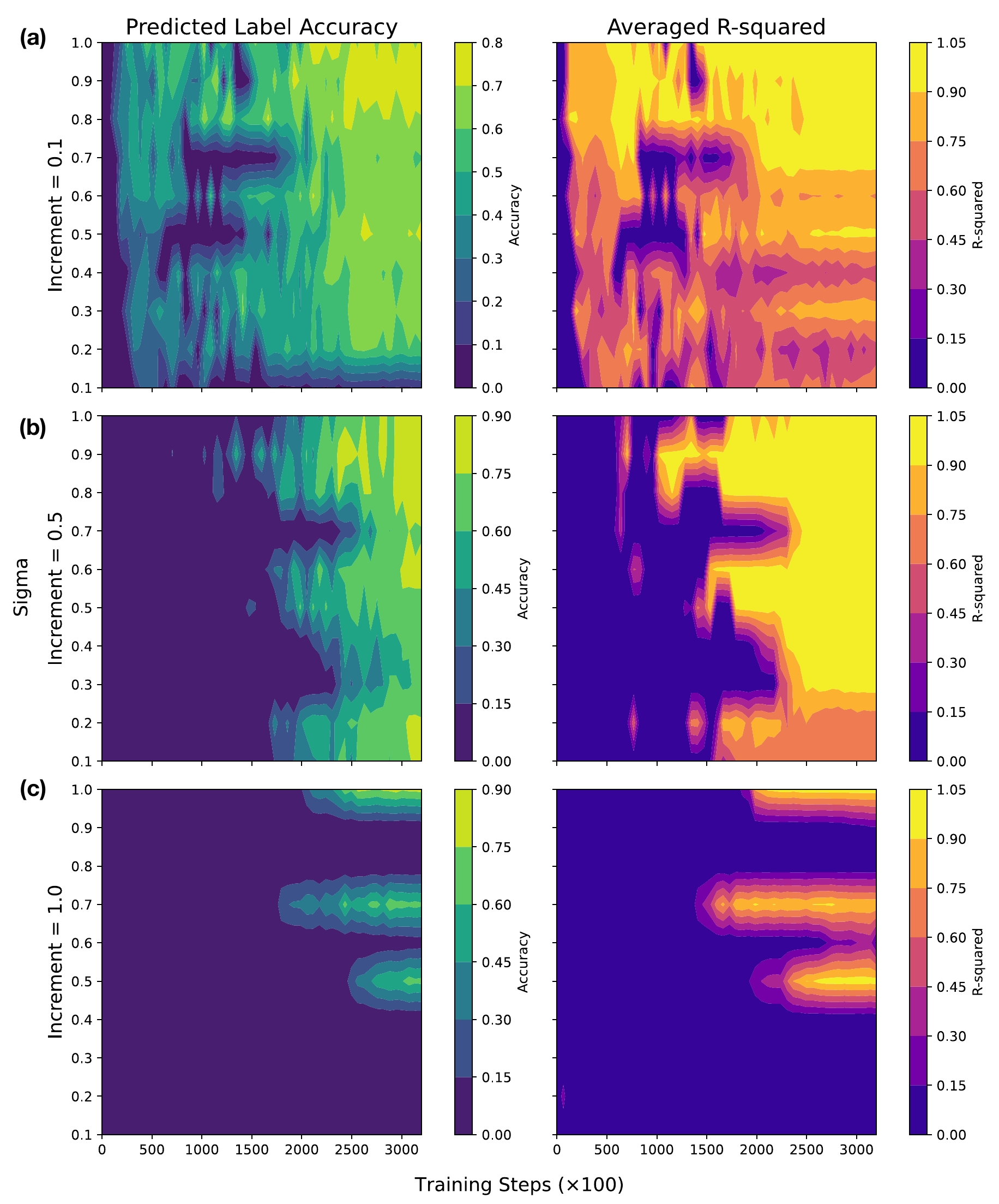}
\end{center}
\caption{\textbf{Phase diagrams of performance metrics as a function of sigmas and training steps for various increments.} \textbf{(a)} Increment = 0.1, \textbf{(b)} Increment = 0.5, \textbf{(c)} Increment = 1.0.}
\label{fig:phase_sigma}
\end{figure}

Previously, in Appendix Sec.~\ref{sec:app_dataset}, we have discussed how spatial information encoded in the datasets varies based on the increment $d$ and the spread $\sigma$. We show in Sec.~\ref{sec:results} that indeed a smaller increment results in a faster rate of convergence leading to a semantically meaningful latent representation. In Fig.~\ref{fig:phase_sigma}, we show the performance metrics as a function of sigma and training steps for three separate increments $d=0.1,0.5,1.0$. There is, however, no strong correlation between model performance and increasing sigma. One possible explanation for this could be the fact that changing sigma only results in a modest change in the dataset (linear to sub-linear in the number of overlapping neighbors), unlike changing increments, which results in a quadratic change in the dataset in addition to more fine-grained embeddings. Moreover, we noticed that for some seeds, some of the sigmas that we have tested do not learn a semantically meaningful manifold. This seed dependence issue is exacerbated with models trained using datasets of bigger increments.

\subsection{Compositional Generalization Performance}
\begin{figure}[h]
\begin{center}
\includegraphics[width=5.0in]{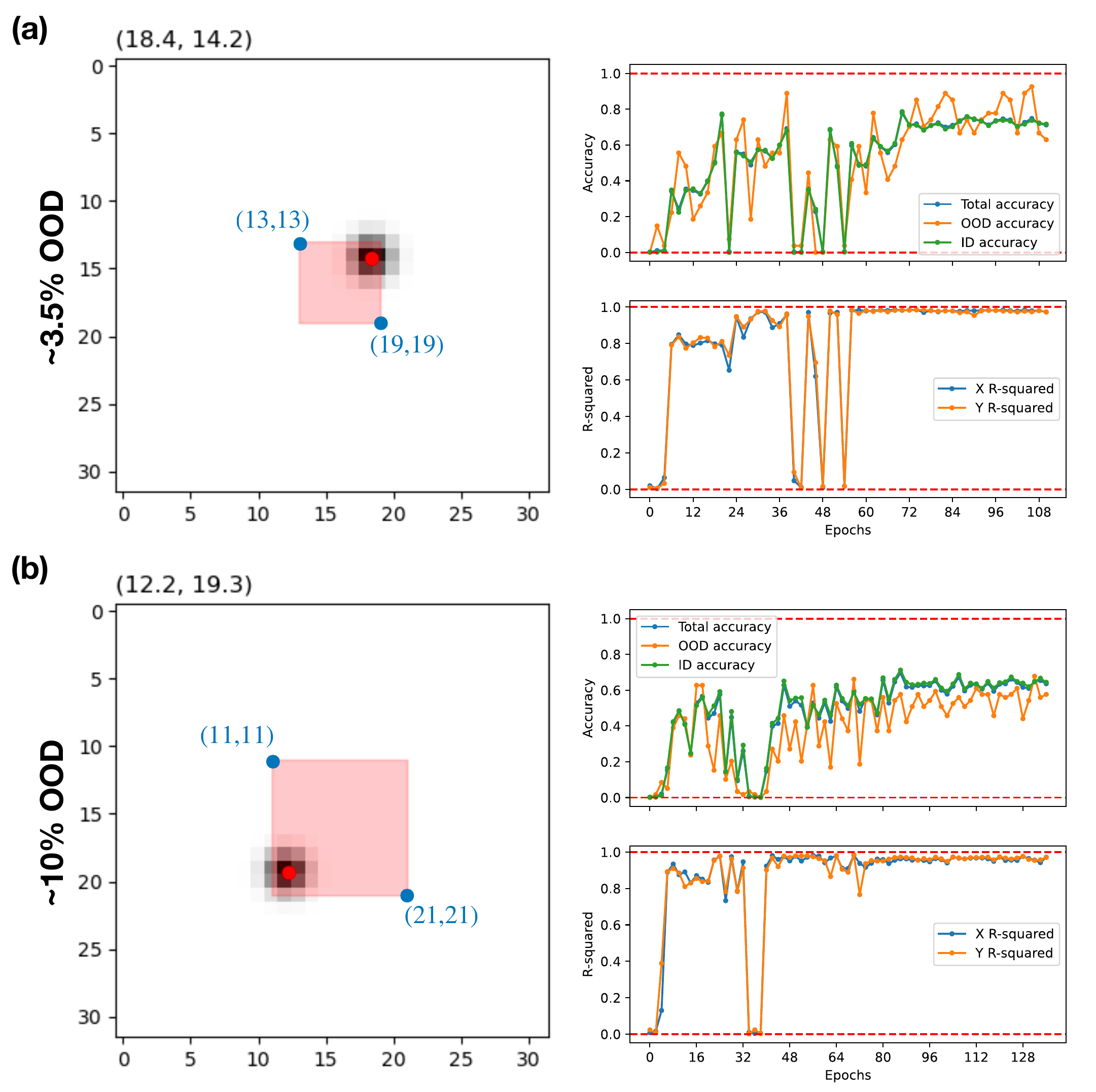}
\end{center}
\caption{\textbf{Performance metrics of models trained under an incomplete set of training data.} Both models are trained using a dataset of $d=0.1$ and $\sigma=1.0$, with \textbf{(a)} $\sim 3.5\%$ and \textbf{(b)} $\sim 10\%$ of the training dataset skipped during training. The training data images with Gaussian bumps centered within the red-shaded regions in the sample OOD images shown on the left are skipped during training.}
\label{fig:compositional_generalization}
\end{figure}

Can the models compositionally generalize well? To answer this question, we train two models under an incomplete training dataset of $d=0.1$ and $\sigma=1.0$, where we deliberately ``poke a hole" in the middle of the data manifold and see if the model can still learn an accurate representation of the dataset. Fig.~\ref{fig:compositional_generalization} shows the performance metrics of the model trained under $\sim 3.5\%$ (a smaller hole) and $\sim 10\%$ (a bigger hole). We note that given sufficient amount of training, both models were able to construct a semantically meaningful 2D representation, with the accuracy of the OOD only slightly worse off in the case where $\sim 10\%$ of the data is skipped as compared to that where only $\sim 3.5\%$ is skipped. These results indicate that the model might have an exceptional interpolation/extrapolation ability or the ability to compositionally generalize. The mechanism of how they generalize OOD warrants further investigation. 

\end{document}

%% file: math_commands.tex
\usepackage{amsmath,amsfonts,bm}

\def\eqref#1{equation~\ref{#1}}

\def\1{\bm{1}}

\DeclareMathAlphabet{\mathsfit}{\encodingdefault}{\sfdefault}{m}{sl}
\SetMathAlphabet{\mathsfit}{bold}{\encodingdefault}{\sfdefault}{bx}{n}














%% file: main.bbl
\begin{thebibliography}{19}
\providecommand{\natexlab}[1]{#1}
\providecommand{\url}[1]{\texttt{#1}}
\expandafter\ifx\csname urlstyle\endcsname\relax
  \providecommand{\doi}[1]{doi: #1}\else
  \providecommand{\doi}{doi: \begingroup \urlstyle{rm}\Url}\fi

\bibitem[Chen et~al.(2021)Chen, Zhang, Zen, Weiss, Norouzi, and Chan]{cond1}
Nanxin Chen, Yu~Zhang, Heiga Zen, Ron~J. Weiss, Mohammad Norouzi, and William Chan.
\newblock Wavegrad: Estimating gradients for waveform generation.
\newblock In \emph{9th International Conference on Learning Representations, {ICLR} 2021, Virtual Event, Austria, May 3-7, 2021}. OpenReview.net, 2021.
\newblock URL \url{https://openreview.net/forum?id=NsMLjcFaO8O}.

\bibitem[Du et~al.(2021)Du, Li, Sharma, Tenenbaum, and Mordatch]{du2021}
Yilun Du, Shuang Li, Yash Sharma, Josh Tenenbaum, and Igor Mordatch.
\newblock Unsupervised learning of compositional energy concepts.
\newblock In Marc'Aurelio Ranzato, Alina Beygelzimer, Yann~N. Dauphin, Percy Liang, and Jennifer~Wortman Vaughan (eds.), \emph{Advances in Neural Information Processing Systems 34: Annual Conference on Neural Information Processing Systems 2021, NeurIPS 2021, December 6-14, 2021, virtual}, pp.\  15608--15620, 2021.
\newblock URL \url{https://proceedings.neurips.cc/paper/2021/hash/838aac83e00e8c5ca0f839c96d6cb3be-Abstract.html}.

\bibitem[Du et~al.(2023)Du, Durkan, Strudel, Tenenbaum, Dieleman, Fergus, Sohl{-}Dickstein, Doucet, and Grathwohl]{du2023}
Yilun Du, Conor Durkan, Robin Strudel, Joshua~B. Tenenbaum, Sander Dieleman, Rob Fergus, Jascha Sohl{-}Dickstein, Arnaud Doucet, and Will~Sussman Grathwohl.
\newblock Reduce, reuse, recycle: Compositional generation with energy-based diffusion models and {MCMC}.
\newblock In Andreas Krause, Emma Brunskill, Kyunghyun Cho, Barbara Engelhardt, Sivan Sabato, and Jonathan Scarlett (eds.), \emph{International Conference on Machine Learning, {ICML} 2023, 23-29 July 2023, Honolulu, Hawaii, {USA}}, volume 202 of \emph{Proceedings of Machine Learning Research}, pp.\  8489--8510. {PMLR}, 2023.
\newblock URL \url{https://proceedings.mlr.press/v202/du23a.html}.

\bibitem[Esmaeili et~al.(2019)Esmaeili, Wu, Jain, Bozkurt, Siddharth, Paige, Brooks, Dy, and van~de Meent]{esmaeili2019}
Babak Esmaeili, Hao Wu, Sarthak Jain, Alican Bozkurt, N.~Siddharth, Brooks Paige, Dana~H. Brooks, Jennifer~G. Dy, and Jan{-}Willem van~de Meent.
\newblock Structured disentangled representations.
\newblock In Kamalika Chaudhuri and Masashi Sugiyama (eds.), \emph{The 22nd International Conference on Artificial Intelligence and Statistics, {AISTATS} 2019, 16-18 April 2019, Naha, Okinawa, Japan}, volume~89 of \emph{Proceedings of Machine Learning Research}, pp.\  2525--2534. {PMLR}, 2019.
\newblock URL \url{http://proceedings.mlr.press/v89/esmaeili19a.html}.

\bibitem[Higgins et~al.(2018)Higgins, Sonnerat, Matthey, Pal, Burgess, Bosnjak, Shanahan, Botvinick, Hassabis, and Lerchner]{higgins2018}
Irina Higgins, Nicolas Sonnerat, Loic Matthey, Arka Pal, Christopher~P. Burgess, Matko Bosnjak, Murray Shanahan, Matthew~M. Botvinick, Demis Hassabis, and Alexander Lerchner.
\newblock {SCAN:} learning hierarchical compositional visual concepts.
\newblock In \emph{6th International Conference on Learning Representations, {ICLR} 2018, Vancouver, BC, Canada, April 30 - May 3, 2018, Conference Track Proceedings}. OpenReview.net, 2018.
\newblock URL \url{https://openreview.net/forum?id=rkN2Il-RZ}.

\bibitem[Ho et~al.(2020)Ho, Jain, and Abbeel]{ho2020denoising}
Jonathan Ho, Ajay Jain, and Pieter Abbeel.
\newblock Denoising diffusion probabilistic models.
\newblock In Hugo Larochelle, Marc'Aurelio Ranzato, Raia Hadsell, Maria{-}Florina Balcan, and Hsuan{-}Tien Lin (eds.), \emph{Advances in Neural Information Processing Systems 33: Annual Conference on Neural Information Processing Systems 2020, NeurIPS 2020, December 6-12, 2020, virtual}, 2020.
\newblock URL \url{https://proceedings.neurips.cc/paper/2020/hash/4c5bcfec8584af0d967f1ab10179ca4b-Abstract.html}.

\bibitem[Liu et~al.(2021)Liu, Li, Du, Tenenbaum, and Torralba]{liu2021}
Nan Liu, Shuang Li, Yilun Du, Josh Tenenbaum, and Antonio Torralba.
\newblock Learning to compose visual relations.
\newblock In Marc'Aurelio Ranzato, Alina Beygelzimer, Yann~N. Dauphin, Percy Liang, and Jennifer~Wortman Vaughan (eds.), \emph{Advances in Neural Information Processing Systems 34: Annual Conference on Neural Information Processing Systems 2021, NeurIPS 2021, December 6-14, 2021, virtual}, pp.\  23166--23178, 2021.
\newblock URL \url{https://proceedings.neurips.cc/paper/2021/hash/c3008b2c6f5370b744850a98a95b73ad-Abstract.html}.

\bibitem[Liu et~al.(2022)Liu, Li, Du, Torralba, and Tenenbaum]{liu2022}
Nan Liu, Shuang Li, Yilun Du, Antonio Torralba, and Joshua~B. Tenenbaum.
\newblock Compositional visual generation with composable diffusion models.
\newblock In Shai Avidan, Gabriel~J. Brostow, Moustapha Ciss{\'{e}}, Giovanni~Maria Farinella, and Tal Hassner (eds.), \emph{Computer Vision - {ECCV} 2022 - 17th European Conference, Tel Aviv, Israel, October 23-27, 2022, Proceedings, Part {XVII}}, volume 13677 of \emph{Lecture Notes in Computer Science}, pp.\  423--439. Springer, 2022.
\newblock \doi{10.1007/978-3-031-19790-1\_26}.
\newblock URL \url{https://doi.org/10.1007/978-3-031-19790-1\_26}.

\bibitem[Liu et~al.(2023)Liu, Du, Li, Tenenbaum, and Torralba]{liu2023}
Nan Liu, Yilun Du, Shuang Li, Joshua~B. Tenenbaum, and Antonio Torralba.
\newblock Unsupervised compositional concepts discovery with text-to-image generative models.
\newblock In \emph{{IEEE/CVF} International Conference on Computer Vision, {ICCV} 2023, Paris, France, October 1-6, 2023}, pp.\  2085--2095. {IEEE}, 2023.
\newblock \doi{10.1109/ICCV51070.2023.00199}.
\newblock URL \url{https://doi.org/10.1109/ICCV51070.2023.00199}.

\bibitem[McInnes \& Healy(2018)McInnes and Healy]{mcinnes2020umap}
Leland McInnes and John Healy.
\newblock {UMAP:} uniform manifold approximation and projection for dimension reduction.
\newblock \emph{CoRR}, abs/1802.03426, 2018.
\newblock URL \url{http://arxiv.org/abs/1802.03426}.

\bibitem[McInnes et~al.(2018)McInnes, Healy, Saul, and Grossberger]{mcinnes2018umap-software}
Leland McInnes, John Healy, Nathaniel Saul, and Lukas Grossberger.
\newblock Umap: Uniform manifold approximation and projection.
\newblock \emph{The Journal of Open Source Software}, 3\penalty0 (29):\penalty0 861, 2018.

\bibitem[Okawa et~al.(2023)Okawa, Lubana, Dick, and Tanaka]{okawa2023}
Maya Okawa, Ekdeep~Singh Lubana, Robert~P. Dick, and Hidenori Tanaka.
\newblock Compositional abilities emerge multiplicatively: Exploring diffusion models on a synthetic task.
\newblock \emph{CoRR}, abs/2310.09336, 2023.
\newblock \doi{10.48550/ARXIV.2310.09336}.
\newblock URL \url{https://doi.org/10.48550/arXiv.2310.09336}.

\bibitem[Ramesh et~al.(2021)Ramesh, Pavlov, Goh, Gray, Voss, Radford, Chen, and Sutskever]{ramesh2021zero}
Aditya Ramesh, Mikhail Pavlov, Gabriel Goh, Scott Gray, Chelsea Voss, Alec Radford, Mark Chen, and Ilya Sutskever.
\newblock Zero-shot text-to-image generation.
\newblock In \emph{International Conference on Machine Learning}, pp.\  8821--8831. PMLR, 2021.

\bibitem[Rombach et~al.(2022)Rombach, Blattmann, Lorenz, Esser, and Ommer]{rombach2022high}
Robin Rombach, Andreas Blattmann, Dominik Lorenz, Patrick Esser, and Bj{\"o}rn Ommer.
\newblock High-resolution image synthesis with latent diffusion models.
\newblock In \emph{Proceedings of the IEEE/CVF conference on computer vision and pattern recognition}, pp.\  10684--10695, 2022.

\bibitem[Saharia et~al.(2022)Saharia, Chan, Saxena, Li, Whang, Denton, Ghasemipour, Gontijo~Lopes, Karagol~Ayan, Salimans, et~al.]{saharia2022photorealistic}
Chitwan Saharia, William Chan, Saurabh Saxena, Lala Li, Jay Whang, Emily~L Denton, Kamyar Ghasemipour, Raphael Gontijo~Lopes, Burcu Karagol~Ayan, Tim Salimans, et~al.
\newblock Photorealistic text-to-image diffusion models with deep language understanding.
\newblock \emph{Advances in Neural Information Processing Systems}, 35:\penalty0 36479--36494, 2022.

\bibitem[Saharia et~al.(2023)Saharia, Ho, Chan, Salimans, Fleet, and Norouzi]{cond2}
Chitwan Saharia, Jonathan Ho, William Chan, Tim Salimans, David~J. Fleet, and Mohammad Norouzi.
\newblock Image super-resolution via iterative refinement.
\newblock \emph{{IEEE} Trans. Pattern Anal. Mach. Intell.}, 45\penalty0 (4):\penalty0 4713--4726, 2023.
\newblock \doi{10.1109/TPAMI.2022.3204461}.
\newblock URL \url{https://doi.org/10.1109/TPAMI.2022.3204461}.

\bibitem[Xu et~al.(2022)Xu, Niethammer, and Raffel]{xu2022}
Zhenlin Xu, Marc Niethammer, and Colin Raffel.
\newblock Compositional generalization in unsupervised compositional representation learning: {A} study on disentanglement and emergent language.
\newblock In Sanmi Koyejo, S.~Mohamed, A.~Agarwal, Danielle Belgrave, K.~Cho, and A.~Oh (eds.), \emph{Advances in Neural Information Processing Systems 35: Annual Conference on Neural Information Processing Systems 2022, NeurIPS 2022, New Orleans, LA, USA, November 28 - December 9, 2022}, 2022.
\newblock URL \url{http://papers.nips.cc/paper\_files/paper/2022/hash/9f9ecbf4062842df17ec3f4ea3ad7f54-Abstract-Conference.html}.

\bibitem[Yang et~al.(2023)Yang, Wang, Lv, and Zheng]{yang2023}
Tao Yang, Yuwang Wang, Yan Lv, and Nanning Zheng.
\newblock Disdiff: Unsupervised disentanglement of diffusion probabilistic models.
\newblock \emph{CoRR}, abs/2301.13721, 2023.
\newblock \doi{10.48550/ARXIV.2301.13721}.
\newblock URL \url{https://doi.org/10.48550/arXiv.2301.13721}.

\bibitem[Zhao et~al.(2018)Zhao, Ren, Yuan, Song, Goodman, and Ermon]{zhao2018}
Shengjia Zhao, Hongyu Ren, Arianna Yuan, Jiaming Song, Noah~D. Goodman, and Stefano Ermon.
\newblock Bias and generalization in deep generative models: An empirical study.
\newblock In Samy Bengio, Hanna~M. Wallach, Hugo Larochelle, Kristen Grauman, Nicol{\`{o}} Cesa{-}Bianchi, and Roman Garnett (eds.), \emph{Advances in Neural Information Processing Systems 31: Annual Conference on Neural Information Processing Systems 2018, NeurIPS 2018, December 3-8, 2018, Montr{\'{e}}al, Canada}, pp.\  10815--10824, 2018.
\newblock URL \url{https://proceedings.neurips.cc/paper/2018/hash/5317b6799188715d5e00a638a4278901-Abstract.html}.

\end{thebibliography}
